\documentclass{article}
\usepackage{amssymb}
\usepackage{amsmath}
\usepackage{subcaption}
\usepackage{graphicx}
\usepackage{makecell}
\usepackage{enumitem}
\usepackage{algorithm}
\usepackage{algorithmic}
\usepackage{arydshln}
\newcolumntype{L}[1]{>{\raggedright\arraybackslash}p{#1}}

\newcommand{\method}{\textsc{EgoZero}}
\newcommand{\website}{\url{https://egozero-robot.github.io}}

\usepackage[preprint]{corl_2025} 

\title{EgoZero: \\ Robot Learning from Smart Glasses}

\author{
  Vincent Liu$^1$\thanks{Correspondence to Vincent Liu: \texttt{vincent.liu15@gmail.com}}
  \qquad
  Ademi Adeniji$^{12*}$
  \qquad
  Haotian Zhan$^{1*}$
  \AND
  Siddhant Haldar$^1$
  \qquad
  Raunaq Bhirangi$^1$
  \qquad
  Pieter Abbeel$^2$
  \qquad
  Lerrel Pinto$^1$
  \AND
  {\normalfont $^1$New York University}
  \qquad
  {\normalfont $^2$UC Berkeley} \\~\\
  {\normalfont $^*$Equal contribution}
}

\begin{document}
\maketitle


\begin{abstract}
Despite recent progress in general purpose robotics, robot policies still lag far behind basic human capabilities in the real world. Humans interact constantly with the physical world, yet this rich data resource remains largely untapped in robot learning. We propose \method{}, a minimal system that learns robust manipulation policies from human demonstrations captured with Project Aria smart glasses, \textbf{and zero robot data}. \method{} enables: (1) extraction of complete, robot-executable actions from in-the-wild, egocentric, human demonstrations, (2) compression of human visual observations into morphology-agnostic state representations, and (3) closed-loop policy learning that generalizes morphologically, spatially, and semantically. We deploy \method{} policies on a gripper Franka Panda robot and demonstrate zero-shot transfer with 70\% success rate over 7 manipulation tasks and only 20 minutes of data collection per task. Our results suggest that in-the-wild human data can serve as a scalable foundation for real-world robot learning — paving the way toward a future of abundant, diverse, and naturalistic training data for robots. Code and videos are available at \website{}.
\end{abstract}

\keywords{Imitation Learning, Robot Learning, Human Data}


\begin{figure}[h]
    \centering
    \includegraphics[width=\linewidth]{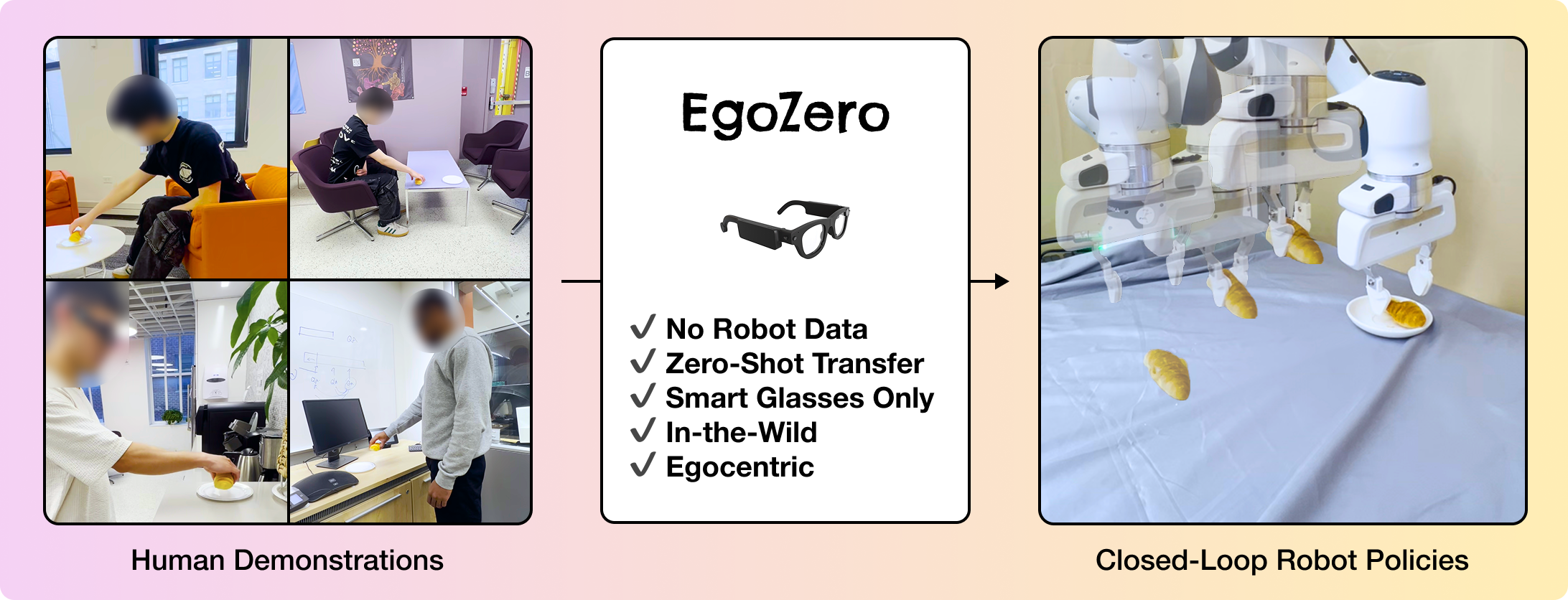}
\end{figure}

\section{Introduction}

Robots face significant challenges in replicating human generality and dexterity in the physical world. While deep learning has fueled progress in domains like language \cite{openai2024gpt4technicalreport, browngpt3}, vision \cite{ramesh2021zeroshottexttoimagegeneration, rombach2022highresolutionimagesynthesislatent, imagenteamgoogle2024imagen3, videoworldsimulators2024, blattmann2023stablevideodiffusionscaling}, speech \cite{evans2024stableaudioopen, wang2023neuralcodeclanguagemodels, radford2022robustspeechrecognitionlargescale}, and complex games \cite{alphazero, openai2019dota2largescale}, these successes rely on internet-scale datasets that are tightly aligned with downstream applications. In robotics, collecting similarly large and diverse datasets that match real-world deployment conditions remains a fundamental bottleneck \cite{embodimentcollaboration2024openxembodimentroboticlearning}.

We argue that the data bottleneck stems not from a shortage of physical labor in the real world, but from the unresolved challenge of effectively capturing and representing human behavior for robot learning. Humans perform a wide range of dexterous tasks in natural environments every day, representing an untapped, renewable source of rich, real-world data. Although recent works have attempted to use human demonstrations as supervision for robot learning, they have limitations to scalability such as additional wearables \cite{wang2024dexcapscalableportablemocap}, robot data \cite{kareer2024egomimicscalingimitationlearning}, multi-camera calibration \cite{haldar2025pointpolicyunifyingobservations}, online fine-tuning \cite{guzey2024bridginghumanrobotdexterity}, low-precision affordance-based policies \cite{bahl2022humantorobotimitationwild, shi2025zeromimicdistillingroboticmanipulation}, or data processing hacks to cross the human-robot morphology gap \cite{wang2024dexcapscalableportablemocap, kareer2024egomimicscalingimitationlearning, lepert2025phantomtrainingrobotsrobots}. Other general vision-based learning approaches pretrain on large multi-robot datasets \cite{embodimentcollaboration2024openxembodimentroboticlearning, khazatsky2024droidlargescaleinthewildrobot}, which produce visual representations that are robust across morphologies present in their training mixes \cite{brohan2022rt, brohan2023rt, black2024pi0visionlanguageactionflowmodel, intelligence2025pi05visionlanguageactionmodelopenworld}, but have yet to show zero-shot transfer purely from human data.

In this work, we tackle the ambitious question: can robots learn zero-shot manipulation skills from only egocentric in-the-wild human data? To answer this, we introduce \method{}: a lightweight framework that enables robots to learn manipulation policies directly from egocentric in-the-wild human demonstrations, captured using only Project Aria smart glasses \cite{engel2023projectarianewtool}. \method{} eliminates the need for teleoperation, calibration, or additional wearables, allowing humans to interact with the world freely while still providing robot supervision. Inspired by \cite{haldar2025pointpolicyunifyingobservations, levy2024p3poprescriptivepointpriors}, \method{} overcomes the morphology gap by representing states and actions as compact sets of points. Point-based representations simultaneously unify human and robot distributions, improve sample efficiency and interpretability of policy learning, and generalize to new visual scenes and morphologies. However, egocentric in-the-wild data collection, does not have access to the multi-camera calibration setup used in \cite{haldar2025pointpolicyunifyingobservations, levy2024p3poprescriptivepointpriors} to accurately compute point representations. Therefore, we introduce methods to accurately derive state and action representations from raw visual and odometric inputs.

We evaluate \method{} by training manipulation policies on human demonstrations recorded by Aria and deploying them on a Franka Panda robot. Our policies achieve an average zero-shot success rate of 70\% across tasks such as grasping, opening, and pick-and-place in unseen real-world environments — without any robot-collected training data. By rethinking the data representation and policy learning stack to be morphology-agnostic from the ground up, \method{} is a step toward building robots that can learn from the vast diversity of real-world human experiences. Our contributions are as follows:
\begin{itemize}[leftmargin=2em]
\item \method{} policies achieve a 70\% zero-shot success rate on our tasks, \textbf{trained only on human data recorded with Project Aria smart glasses}. \method{}, to our knowledge, represents the first approach that successfully transfers in-the-wild, human data into closed-loop policies with no robot data.
\item \method{} policies exhibit strong zero-shot generalization properties with only 100 training demonstrations (20 minutes of data collection), demonstrating the robustness, transferability, and data efficiency of learning from unified 3D state-action representations.
\item \method{} achieves high success rate when evaluated on new camera viewpoints, spatial configurations, and object instances that are often completely out-of-distribution — validating our proposed method of extracting accurate 3D representations from objects when accurate depth measurements are not available.
\end{itemize}

\section{Related Work}

\textbf{Imitation learning.}
Imitation learning has emerged as a powerful paradigm in robotics, enabling robots to acquire complex skills by learning directly from real-world demonstrations \cite{mandlekar2021what}. By observing and replicating expert behavior, robots can bypass the need for hand-engineered solutions to manipulation tasks, making this approach particularly conducive to domains with high-dimensional state and action spaces \cite{mandlekar2019scaling, jang2022}. Teleoperation is one of the most widely used methods for imitation learning from real-world data collection and has been extensively studied in the robotics literature. In this approach, a human teleoperator commands a robot to complete a desired task, recording the robot's states and actions in the process. The collected data is then used to train a policy that predicts actions from states via supervised learning \cite{argall2009survey, hussein2017imitation, zhao2023learning, zhao2024alohaunleashedsimplerecipe, wu2024gellogenerallowcostintuitive}.

\textbf{Learning from human motion.} Because teleoperation is difficult to scale due to its hardware requirements, learning manipulation directly from humans has become a growing area of interest. Prior work has explored mapping human grasps to robot manipulators using vision-based representations like the “contact web” \cite{kang1994toward}, and more recently, has introduced semantic constraints to encode the implicit common sense required for household tasks \cite{ikeuchi2023semantic}. Other methods to capture human proprioception include ``inside-out'' motion capture systems such as VR headsets and dongles, which do not use external sensing devices \cite{metaquest, visionpro} and are bulky, tethered, and susceptible to occlusion. SLAM-based wearable camera systems \cite{wang2024dexcapscalableportablemocap} and VR wrist trackers such as the SteamVR wrist trackers \cite{steamvr} are vision-based and do not require external transmitters for localization, but can drift and become inaccurate. Self-tracking vision methods require extensive calibration and mapping of each environment a priori \cite{wang2024dexcapscalableportablemocap}. For capturing local information such as finger movements, motion capture gloves such as Rokoko and Manus Metagloves are highly accurate \cite{mannam2023, xsens, manus, rokoko}. These gloves use resistive strain sensing, capacitative sensing, and electromagnetic field sensing to track precise finger information in the local hand frame.

\textbf{Learning from egocentric video.} Because of the accessibility of video, several recent works try to learn and extract hand data from egocentric videos of humans. Datasets such as \cite{Shan20, grauman2022ego4d, damen2018scaling, Goyal_2017_ICCV, chao2021dexycb} represent large-scale efforts to collect egocentric videos of humans interacting with objects in diverse real-world scenes. \cite{levy2024p3poprescriptivepointpriors, haldar2025pointpolicyunifyingobservations} use point-based representations to unify human video and robot training data, while \cite{lepert2025phantomtrainingrobotsrobots} modifies human videos with image editing models to create robot training data. \cite{kareer2024egomimicscalingimitationlearning, qin2022dexmv} propose hardware solutions such as smart glasses and multi-camera data collection platforms to collect dexterous hand video datasets, while \cite{bahl2022humantorobotimitationwild, papagiannis2024rxretrievalexecutioneveryday, bahl2023affordances, singh2024handobjectinteractionpretrainingvideos, shi2025zeromimicdistillingroboticmanipulation} introduce methods for extracting control-based affordances for manipulation from vision. Many of the approaches that estimate hand pose information from one or more camera inputs are facilitated by hand-pose estimation models such as \cite{pavlakos2023reconstructing, zhang2019, baek2019, boukhayma2019}. These models are trained with imitation learning to predict hand keypoints \cite{romero2022} from monocular visual input. Although effective in many simple domains, these models are brittle to occlusions, temporally inconsistent, and lack robustness to background distractors.

\section{\method{}}

In this section, we describe \method{}, a system for collecting in-the-wild egocentric human data and training morphology-agnostic robot manipulation policies. 

\subsection{Human-Robot Domain Unification}

\textbf{Project Aria smart glasses.} The Project Aria smart glasses come with several sensors, an SDK, and additional Machine Perception Services (MPS) \cite{engel2023projectarianewtool}. We use the fisheye RGB camera and 2 SLAM cameras for data capture. We obtain accurate online 6DoF hand poses, camera intrinsics, and camera extrinsics from MPS. We record demonstrations of RGB images, 6DoF palm poses, and 6DoF camera extrinsics, which we denote for a timestep $t$ as $(I_t, H_t, T_t)$, respectively. We linearize $I_t$ as a 1408x1408 RGB image with known camera projection function $\mathcal{P}$ and $H_t, T_t \in SE(3)$ are homogeneous transformation matrices representing the hand pose in camera frame and the camera frame in world frame, respectively.

Traditionally, $\mathcal{S}$ represents the robot's space of visual states and $\mathcal{A}$ represents the robot's native executable actions. Similar to \cite{haldar2025pointpolicyunifyingobservations}, we define the morphology-agnostic state and action spaces $\tilde{\mathcal{S}}$ and $\tilde{\mathcal{A}}$, respectively, in egocentric frame. In this section, we describe how to extract $\tilde{\mathcal{S}} \times \tilde{\mathcal{A}}$ from a demonstration $\{(I_t, H_t, T_t)\}_{t=1}^L$.

\textbf{Unified action space.} We define $\tilde{\mathcal{A}}$ as the concatenated space of 3D end-effector egocentric coordinates and gripper closures \cite{wang2024dexcapscalableportablemocap}. Aria only provides $H_t$, which contains no end-effector information except for hand pose \cite{kareer2024egomimicscalingimitationlearning}. We use HaMeR \cite{pavlakos2023reconstructing} to compute the 21-keypoint egocentric hand model, $h_t \in \mathbb{R}^{21 \times 3}$. Though HaMeR's end-effector predictions in camera frame are inaccurate, its predictions localized in hand frame are more reliable. Therefore, we compose local hand deformation from HaMeR with egocentric hand information from Aria. First, we construct HaMeR's palm in camera frame as $\hat{H}_t \in SE(3)$: the translation is the centroid of the ThumbCMC, IndexMCP, and MiddleMCP points; the rotation is the basis constructed by the Wrist-MiddleMCP and IndexMCP-MiddleMCP vectors. We then use $H_t$ to correct $\hat{H}_t$ in egocentric frame through the palm frames. Finally, we project the corrected hand pose $H_t^{-1}\hat{H}_t$ into the first frame \cite{wang2024dexcapscalableportablemocap, kareer2024egomimicscalingimitationlearning}. This can be represented as a single chain of homogeneous transformations
\begin{equation} \label{eq:action_space}
    \tilde{h}_t = T_0^{-1}T_tH_t^{-1}\hat{H}_th_t
\end{equation}

To detect grasps, we threshold the Euclidean distance between the thumb and index coordinates. Our final action is the concatenated vector of thumb and index coordinates and gripper closure.

\begin{figure}[t]
    \centering
    \includegraphics[width=\textwidth]{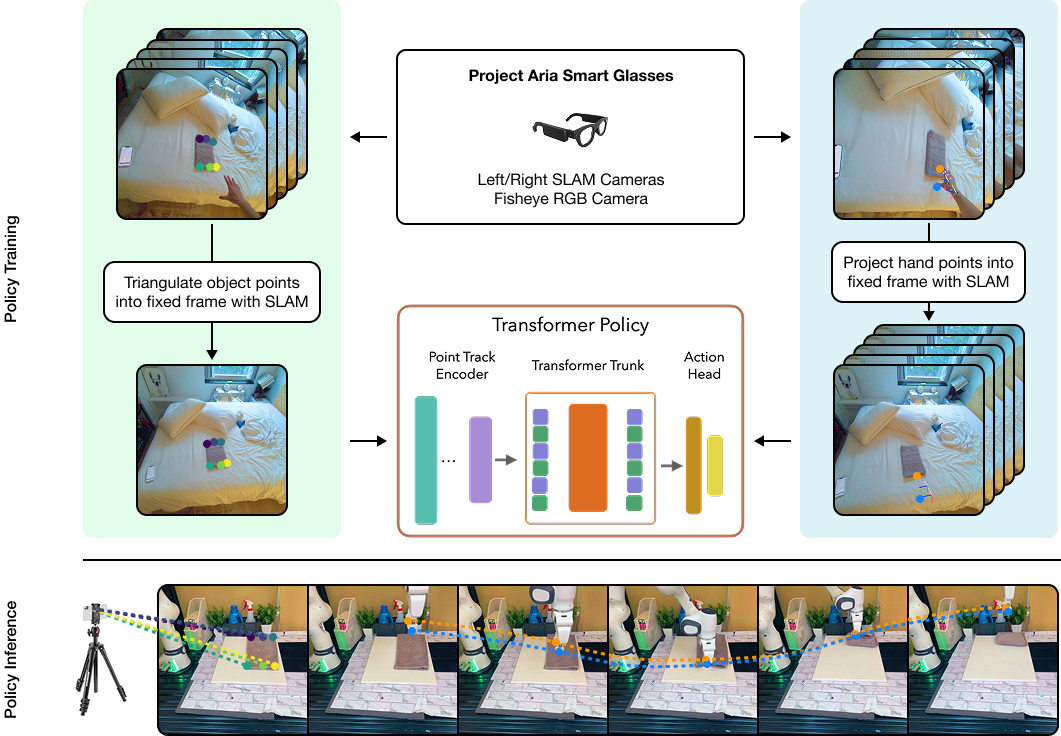}
    \caption{\method{} trains policies in a unified state-action space defined as egocentric 3D points. Unlike previous methods which leverage multi-camera calibration and depth sensors, \method{} localizes object points via triangulation over the camera trajectory, and computes action points via Aria MPS hand pose and a hand estimation model. These points supervise a closed-loop Transformer policy, which is rolled out on unprojected points from an iPhone during inference.}
    \label{fig:architecture}
\end{figure}

\textbf{Unified state space.} We define $\tilde{\mathcal{S}}$ as the concatenated space of egocentric object point sets and robot end-effector actions. Extracting point representations of objects requires either triangulation from multiple cameras or unprojection with depth, but the Project Aria glasses provide neither\footnote{Though there are 3 visual cameras (1 RGB, 2 SLAM), they have little field-of-view overlap, making stereo triangulation unreliable \url{https://github.com/facebookresearch/projectaria_tools/issues/64}.}. Furthermore, monocular metric depth models are inconsistent and inaccurate even with grounding, which we show in Appendix \ref{appendix:monocular}. Instead, we rely on Aria's accurate SLAM extrinsics and CoTracker3 \cite{karaev2024cotracker3simplerbetterpoint} to triangulate 2D points over the demonstration trajectory. This makes the following assumptions: (1) the object is stationary pre-grasp, (2) there is enough camera movement, and (3) the environment is not stochastic. As such, the object state is static for the entire demonstration.

We first label a set of 2D points \cite{levy2024p3poprescriptivepointpriors, haldar2025pointpolicyunifyingobservations}. For each expert-labeled point, we use Grounding DINO \cite{liu2024groundingdinomarryingdino} and DIFT \cite{tang2023emergentcorrespondenceimagediffusion} to map its UV coordinates onto the start frame, and track these points with CoTracker3 \cite{karaev2024cotracker3simplerbetterpoint} to obtain a trajectory of $(T_t, u_t)$ pairs where $u_t \in \mathbb{R}^2$ and $T_t \in SE(3)$ is the camera pose in world frame. We wish to solve for the $\mathbf{q}^*$ in the first frame ($t=0$) that minimizes the pixel reprojection error in each frame. First, we find a set of inlier frames $\mathcal{I}$ via epipolar geometric consistency and RANSAC triangulation. CoTracker3 oftentimes predicts points that lag behind camera movement, giving the impression that a point is further in space than it actually is. To account for this ``stickiness,'' we add a soft depth penalty to prefer closer solutions when there are multiple points in the cone of solutions that minimize reprojection error. Therefore, we solve
\begin{equation}
    \mathbf{q}^* = \arg\min_\mathbf{q}\sum_{i\in\mathcal{I}} \left|\left|u_i - \mathcal{P}\left(T_0^{-1}T_i\mathbf{q}\right)\right|\right|_\rho + \lambda \mathbf{q}_z
\end{equation}
where $||\cdot||\rho$ is the Huber loss, $\mathcal{P}$ is the camera projection function, and $\lambda$ is the depth penalty weight. In practice, $(T_t, u_t)$ are accurate, so $\mathcal{I}$ contains most of the frames and optimization converges strongly to a mean inlier reprojection error of 2-4 pixels per demonstration. Finally, we order and concatenate all triangulated points to represent the object state, $\tilde{s}$. We provide comprehensive mathematical equations for this procedure in Appendix \ref{appendix:triangulation}.

\subsection{Learning a Robot Policy on Human Data}

\textbf{Policy learning.} We collect $N$ human demonstrations and process them into a dataset $\mathcal{D} = \{(\tilde{s}^{(i)}, \tilde{a}^{(i)})\}_{i=1}^N$. We train a closed-loop Transformer policy \cite{levy2024p3poprescriptivepointpriors} $\pi_\theta: \tilde{\mathcal{S}} \mapsto \tilde{\mathcal{A}}$ with behavior cloning over $\mathcal{D}$. We model the policy's predictions as the mean of a normal distribution and train it to minimize the negative log likelihood function
\begin{equation}
    \theta = \arg\min_\theta \mathbb{E}_{(\tilde{s}, \tilde{a}) \sim \mathcal{D}} \left[ \frac{||\pi_\theta(\tilde{s}) - \tilde{a}||^2}{2\sigma^2} \right]
\end{equation}

where $\sigma = 0.1$ \cite{haldar2024bakuefficienttransformermultitask, levy2024p3poprescriptivepointpriors}. We augment the policy with a history buffer input and temporally aggregated action chunking \cite{haldar2024bakuefficienttransformermultitask, levy2024p3poprescriptivepointpriors}. We randomly inject noise into the object points and apply random 3D transformations to the states and actions of each training episode \cite{wang2024dexcapscalableportablemocap}, which we show is necessary for in-the-wild transfer in Section \ref{ablations}. To do so, we sample random rotations $R \sim \mathcal{U}(-\pi/6, +\pi/6)$ radians and translations $t \sim \mathcal{U}(-0.5, +0.5)$ meters. We remove stationary points by throwing out consecutive points whose Euclidean distance is less than 1cm, which is necessary to disambiguate the association between proprioceptive position and grasp closure. For longer tasks, we subsample the demonstrations by a factor of 2. To discard noisy training examples from DIFT failures, we discard demonstrations whose object points are more than 1 median absolute deviation distance from the closest human fingertip point.

\textbf{Policy inference.}
In inference, we initialize the robot state 30 centimeters above the middle of its workspace. We use Grounding DINO and DIFT to crop and map the expert-labeled UV coordinates onto the start frame. We use an iPhone to represent the stationary egocentric view since it allows us to unproject points into 3D with accurate depth. To map the policy's 3D predictions into robot frame, we calibrate the iPhone-to-robot transform once at the start of inference. We binarize the model's gripper predictions at 0 to produce gripper actions in $\{-1, 1\}$. In our experiments, we use a Franka Panda gripper robot, whose controller produces robot-executable actions via the inverse kinematics mapping $\tilde{\mathcal{A}} \mapsto \mathcal{A}$.

\section{Experiments}

In this section, we compare \method{} with baselines adapted from related works and ablate some of \method{}'s core components. From these comparisons, we demonstrate how our specific design choices make zero-shot in-the-wild transfer possible. We also explore the generalization properties that emerge from \method{}'s unified state-action representation space.

\begin{figure}[t]
  \centering
  \begin{subfigure}[t]{\textwidth}
    \centering
    \includegraphics[width=0.23\textwidth]{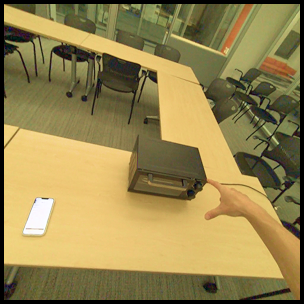}
    \hspace{1mm}
    \includegraphics[width=0.23\textwidth]{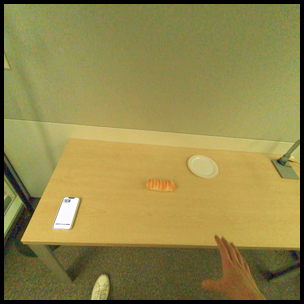}
    \hspace{1mm}
    \includegraphics[width=0.23\textwidth]{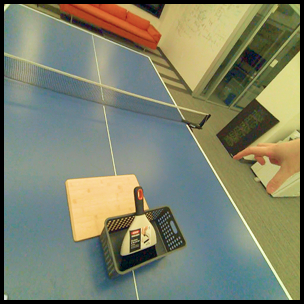}
    \hspace{1mm}
    \includegraphics[width=0.23\textwidth]{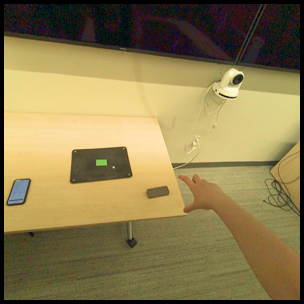}
  \end{subfigure}

  \vspace{2mm} 

  \begin{subfigure}[t]{\textwidth}
    \centering
    \hspace*{.125\%} 
    \includegraphics[width=0.23\textwidth]{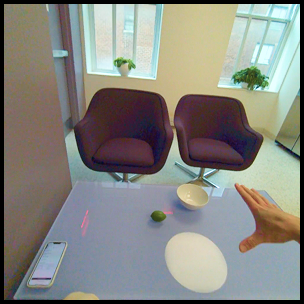}
    \hspace{1mm}
    \includegraphics[width=0.23\textwidth]{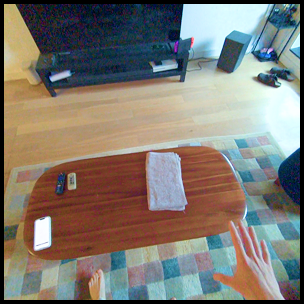}
    \hspace{1mm}
    \includegraphics[width=0.23\textwidth]{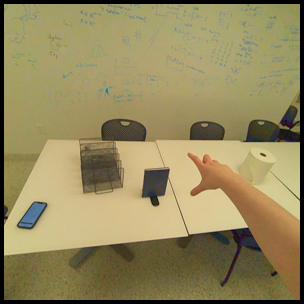}
  \end{subfigure}
  \caption{Our 7 tasks. Top: open oven door, put bread on plate, sweep board with broom, erase board. Bottom: sort fruit, fold towel, and insert book in shelf. See Appendix \ref{appendix:tasks} for full trajectories.}
\end{figure}

\subsection{Experimental Setup}

We evaluate \method{} on a Franka Panda gripper robot. We use an iPhone to represent the egocentric point of view and calibrate this to the robot's frame once per evaluation via an Aruco tag, which we cover during policy inference. We collect 100 demonstrations per task, varying the environment and object positions. \textbf{We collect zero data in our inference-time environment.} We evaluate our method on the following manipulation tasks:
\begin{itemize}[leftmargin=2em]
    \item \textit{Open oven door.} The robot arm grasps and pulls down the handle of an oven door. The position of the oven is varied for each evaluation.
    \item \textit{Put bread on plate.} The robot arm picks up a deformable slice of bread from the table and puts it on the plate. The positions of the bread are varied for each evaluation.
    \item \textit{Sweep board with broom.} The robot arm picks up a mini broom from the basket and sweeps a wooden board. The positions of the broom, basket, and board are varied for each evaluation.
    \item \textit{Erase board.} The robot arm picks up a whiteboard eraser from the table and erases a whiteboard with it. The positions of the eraser and board are varied for each evaluation.
    \item \textit{Sort fruit into bowl.} The robot arm is prompted to pick up one of a lemon, lime, and tangerine, and drop it into a bowl. The positions of the fruits and bowl are varied for each evaluation.
    \item \textit{Fold towel.} The robot arm lifts one end of the towel (closest to the camera) and folds it onto the other end of the towel. The position of the towel is varied for each evaluation.
    \item \textit{Insert book in shelf.} The robot arm picks up a book and inserts it into a shelf. The positions of the book and shelf are varied for each evaluation.
\end{itemize}

\subsection{Baselines}
\label{baselines}

In this section, we demonstrate why our specific formulation of policy learning enables zero-shot transfer from in-the-wild human behaviors. Because no prior work operates under the same assumptions as ours — learning a closed-loop policy in-the-wild, untethered, without robot data, from only smart glasses — we adapt some ideas inspired by past works to our setting.

\textbf{Learning from images.} We implement a variation of Baku \cite{haldar2024bakuefficienttransformermultitask} that predicts actions in our unified action space from image inputs. Due to the large differences in visual distributions between humans and robots, it is difficult to learn a closed-loop policy from human video with zero-shot robot transfer. \cite{kareer2024egomimicscalingimitationlearning} only shows experiments using human video from Aria glasses as supplementary to robot data, requiring careful renormalization of the human data distribution. Furthermore, Aria's fisheye lens exacerbates this problem by warping the 2D-3D correspondence non-uniformly across space and time. Learning 3D distributions from 2D context clues becomes more reliable with abundant visual data produced by similar robot and camera distributions \cite{kim2024openvlaopensourcevisionlanguageactionmodel, black2024pi0visionlanguageactionflowmodel, intelligence2025pi05visionlanguageactionmodelopenworld}.

\textbf{Learning from affordances.} \cite{bahl2022humantorobotimitationwild, shi2025zeromimicdistillingroboticmanipulation} explores learning from egocentric human video data without robot data in affordance-based settings. Typically, this is done by relying on an open-loop trajectory generated by a pretrained grasp model. We ablate our closed-loop formulation by predicting proprioceptive landmarks similar to \cite{bahl2022humantorobotimitationwild} — specifically, the initial and final grasp, executing a linear trajectory between them during inference. Although policy learning from affordances is simple with 3D representations, it fails on tasks that require complex nonlinear motions, such as our ``put bread in plate'' and ``erase board '' tasks. When deployed on the robot, these policies exhibit incorrect behavior: the robot attempts to drag the bread onto the plate and pushes the board with the eraser. In other partially successful tasks, the policy fails by generating trajectories that are too simple, often bumping other objects during execution. These failures demonstrate that closed-loop policies are necessary to learn complex motions with greater precision, even when the object state is not tracked.

\newcolumntype{L}[1]{>{\raggedright\arraybackslash}m{#1}}
\renewcommand{\arraystretch}{1.5}
\begin{table}[t]
\centering
\begin{tabular}{L{4.35cm}ccccccc}
\hline
\makecell{\rule{0pt}{2.5ex}\textbf{Method}} & 
\makecell{\rule{0pt}{2.5ex}\textbf{Open}\\\textbf{oven}} & 
\makecell{\rule{0pt}{2.5ex}\textbf{Pick}\\\textbf{bread}} & 
\makecell{\rule{0pt}{2.5ex}\textbf{Sweep}\\\textbf{broom}} & 
\makecell{\rule{0pt}{2.5ex}\textbf{Erase}\\\textbf{board}} & 
\makecell{\rule{0pt}{2.5ex}\textbf{Sort}\\\textbf{fruit}} & 
\makecell{\rule{0pt}{2.5ex}\textbf{Fold}\\\textbf{towel}} & 
\makecell{\rule{0pt}{2.5ex}\textbf{Insert}\\{\textbf{book}}} \\
\hline
From vision \cite{haldar2024bakuefficienttransformermultitask} & 0/15 & 0/15 & 0/15 & 0/15 & 0/15 & 0/15 & 0/15 \\
From affordances \cite{bahl2022humantorobotimitationwild} & 12/15 & 0/15 & 0/15 & 0/15 & 7/15 & 10/15 & 5/15 \\
\hdashline
\method{} - 3D augmentations & 0/15 & 0/15 & 0/15 & 0/15 & 0/15 & 0/15 & 0/15 \\
\method{} - triangulated depth & 0/15 & 0/15 & 0/15 & 0/15 & 0/15 & 0/15 & 0/15 \\
\hdashline
\textbf{\method{}} & \textbf{13/15} & \textbf{11/15} & \textbf{9/15} & \textbf{11/15} & \textbf{11/15} & \textbf{10/15} & \textbf{9/15} \\
\hline
\end{tabular}
\vspace{5pt}
\caption{Success rates for all baselines and ablations. All models were trained on the same 100 demonstrations per task, and evaluated on zero-shot object poses (unseen from training), cameras (iPhone vs Aria), and environment (robot workspace vs in-the-wild). Because of limited prior work in our exact zero-shot in-the-wild setting, we cite the closest work for each baseline.}
\label{tab:combined}
\end{table}

\subsection{Ablations}
\label{ablations}

In this section, we explore the critical design components that make zero-shot transfer from in-the-wild human data possible. Through our ablative experiments, we argue that the fully egocentric framework necessitates some aspects of policy learning that were not important in more constrained settings.

\begin{figure}[t]
  \centering
  \begin{subfigure}[t]{\textwidth}
    \includegraphics[width=\textwidth]{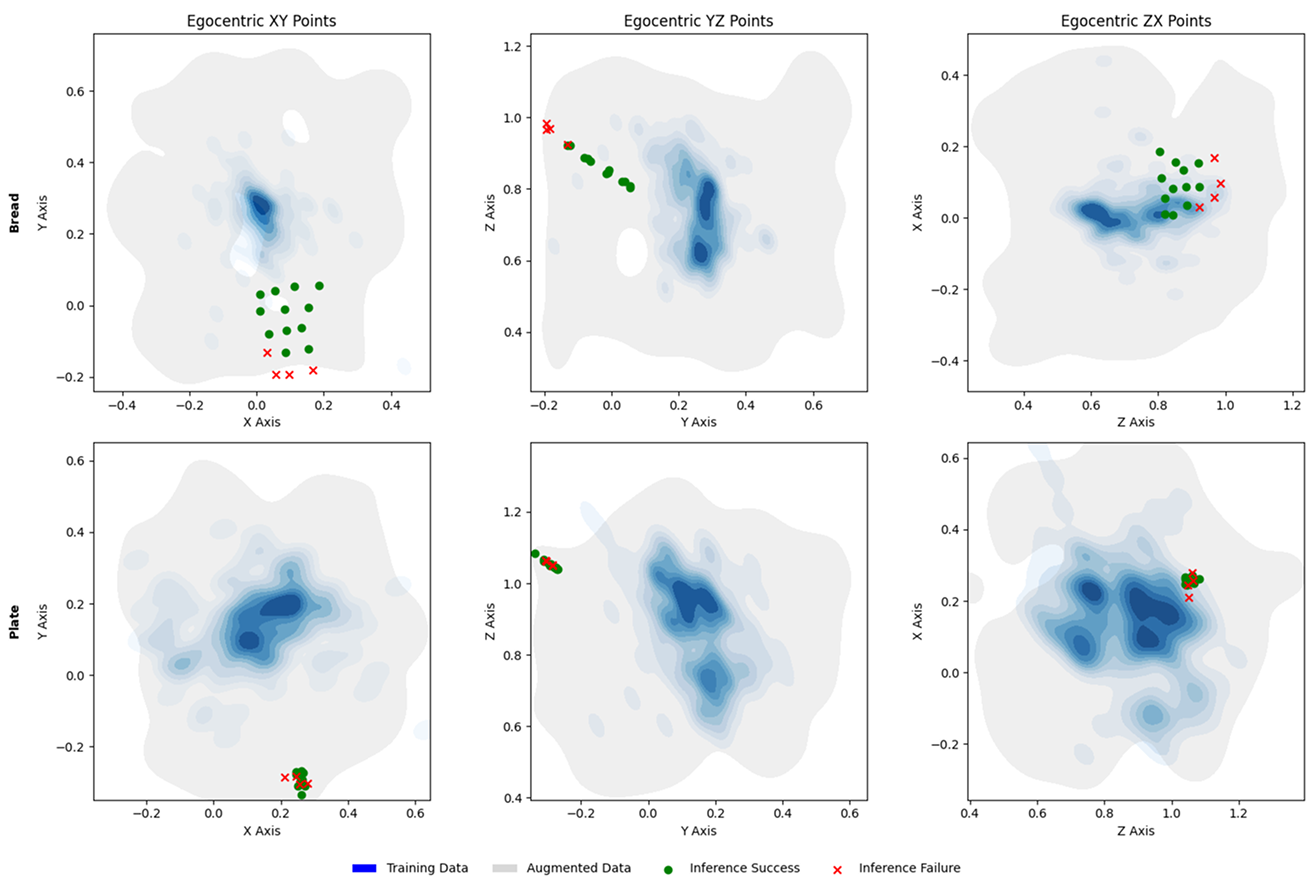}
  \end{subfigure}
  \caption{Distribution of bread keypoints for ``Put bread in plate'' task. The columns are projections of the 3D space onto each 2D plane. The policy generalizes to object poses far outside of its training volume and begins to fail when the objects are near the limits of its augmented volume.}
  \label{fig:3d}
\end{figure}

\textbf{3D augmentations.} Although 3D augmentations have been explored before \cite{wang2024dexcapscalableportablemocap}, we show that they are indeed necessary for zero-shot in-the-wild transfer. In the unified 3D state-action space, the policy learns a dense 3D-to-3D mapping \cite{haldar2025pointpolicyunifyingobservations}. Without 3D augmentations, the policy learns a smaller and sparser 3D-to-3D mapping volume. As a result, the policy does not interpolate between 3D positions as well and is less robust to new positions. Therefore, it is often out-of-distribution when given a new egocentric view. We demonstrate that, when trained with 3D augmentations, our policies generalize to object configurations that are many standard deviations outside of the volume of their training data. Although our policy learning framework is similar to \cite{levy2024p3poprescriptivepointpriors, haldar2025pointpolicyunifyingobservations}, these works do not need 3D augmentations to show good success rates, implying that learning robust policies on egocentric data introduces extra complexity in learning generalizable representations. We visualize the training and inference distributions of object points in Figure \ref{fig:3d}.

\textbf{Monocular depth estimation.} The Aria glasses do not provide a way of extracting ground truth depth information: (1) it cannot triangulate objects reliably since the overlapping field-of-view between all cameras is narrow; (2) it does not have any built-in lidar or depth sensors. Therefore, we localize the object via triangulation over the camera trajectory to obtain its 3D information. To show that monocular metric depth models are not a viable option, we ablate our triangulation method with unprojection from a metric depth model \cite{Bochkovskii2024:arxiv}. We observe that the best metric depth models, even when grounded with many Aruco tags in the scene, produce depth measurements of $>$5cm error. This suggests that the depth maps are warped unevenly, potentially by the distortion caused by Aria's fisheye. All policies trained with estimated depth fail unequivocally. We describe our grounding method in Appendix \ref{appendix:monocular}.

\subsection{Zero-shot generalization}

\textbf{Object pose generalization.} In both data collection and robots evaluation, we vary the poses of the objects. If there are multiple objects, we also vary their locations relative to each other. We observe that the use of correspondence with 3D state representations encodes the pose of the object \cite{levy2024p3poprescriptivepointpriors, haldar2025pointpolicyunifyingobservations} and allows our policies to generalize from in-the-wild data. We notice that there is much more spatial diversity in our human demonstrations than what the robot can access in its workspace. This diversity, combined with 3D augmentations, regularizes the policy to learn a more general solution across a larger 3D volume, which enables zero-shot transfer to the robot. We constrain the diversity of object poses to represent what a human will realistically manipulate (i.e. the oven door is visible to the camera).

\textbf{Object semantic generalization.} Following \cite{levy2024p3poprescriptivepointpriors, haldar2025pointpolicyunifyingobservations}, we also demonstrate that 3D representations allow for zero-shot object category generalization. Because our training and inference images are so different (Aria fisheye vs iPhone pinhole), we introduce Grounding DINO to crop images to improve DIFT's success rate; this is not something that \cite{levy2024p3poprescriptivepointpriors, haldar2025pointpolicyunifyingobservations} implement because their cameras and backgrounds are identical between training and inference. Because Grounding DINO is language-conditioned, we simply prompt it with the object category (i.e. ``a toaster oven.'') to allow it to generalize to entirely new object instances. This ensembling of pretrained models compresses visual diversity into geometric abstractions that allow \method{} to generalize across visual distributions in the egocentric setting.

\begin{figure}[t]
  \centering
  \begin{subfigure}[b]{\textwidth}
    \includegraphics[width=\textwidth]{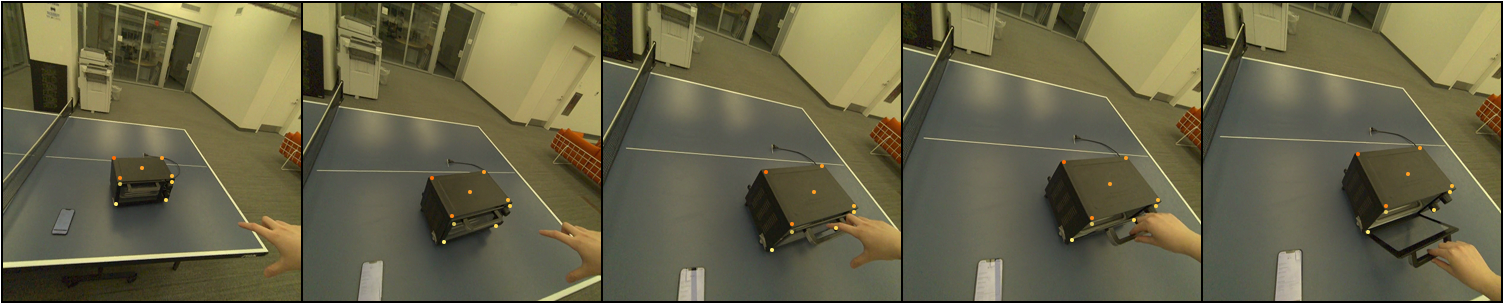}
  \end{subfigure}
  \hfill
  \begin{subfigure}[b]{\textwidth}
    \includegraphics[width=\textwidth]{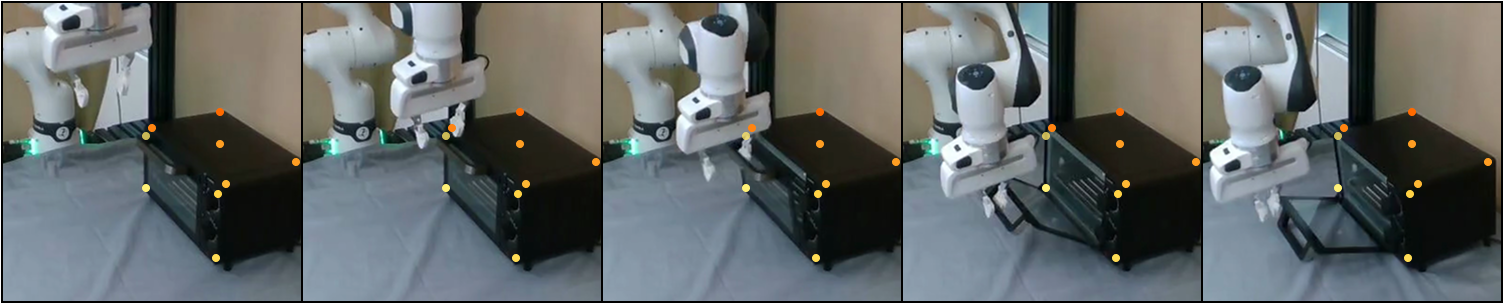}
  \end{subfigure}
  \hfill
  \begin{subfigure}[b]{\textwidth}
    \includegraphics[width=\textwidth]{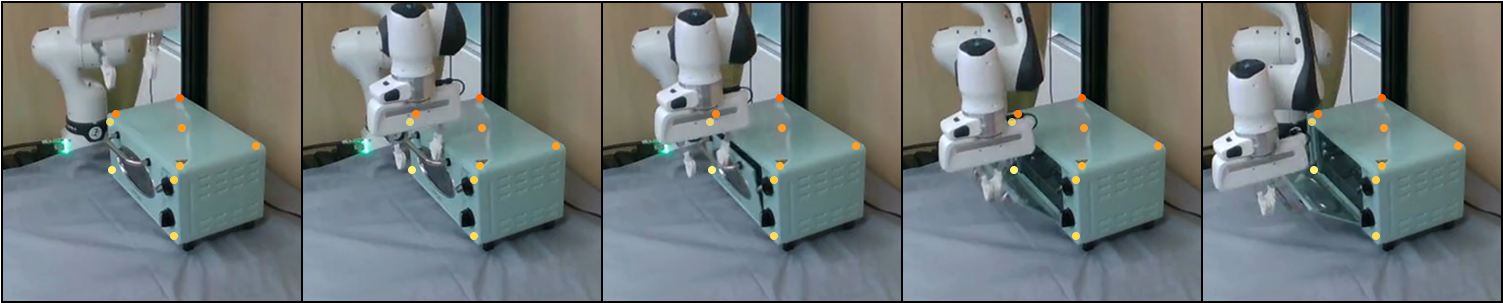}
  \end{subfigure}
  \caption{Object semantic generalization. Human demonstrations are done with only black ovens (top). The policy transfers zero-shot to the robot with the same oven (middle) and also generalizes to a new oven instance (bottom). The points are color-coded to represent the correspondence.}
\end{figure}

\textbf{Camera generalization.} One of the biggest limiting factors of vision-based policies is that learning invariance to small changes in individual pixels is data intensive. For a policy to generalize to novel viewing angles, distances, and cameras, it must be trained on a large amount of data from similar visual distributions. For example, \cite{black2024pi0visionlanguageactionflowmodel} is trained on 10k+ hours of cross-embodiment data, but its zero-shot performance is significantly lower when the inference camera (and end-effector) is different from the one used to collect its robot training data. To navigate this issue, \cite{kareer2024egomimicscalingimitationlearning} uses Aria glasses for human data collection, robot data collection, and policy inference, but still require several hours of both human and robot data and careful renormalization to reach good success rates. Because \method{} learns policies from 3D point sets, \method{} is completely camera-agnostic. We demonstrate this in all our experiments by using an iPhone in inference.

\textbf{Human-scale generalization.} For each task, we collect data in 2-3 different environments, on tabletops of different heights, with various background distractors, with multiple unique demonstrators. We perform our demonstrations moving around, standing still, and sitting down. The variance in human demonstrators provides added diversity in the training data. These differences in height and grasp are still encoded in the same unified representation space.

\subsection{Limitations}
\textbf{Limitations of 3D representations.} The largest source of error during inference comes from the correspondence model DIFT \cite{tang2023emergentcorrespondenceimagediffusion}. Correspondence encodes pose by ordering the state space, making policy learning sample efficient \cite{levy2024p3poprescriptivepointpriors, haldar2025pointpolicyunifyingobservations}. At larger data scale, pose information can be learned directly from dense unordered geometric information (i.e. using grounded segmentation models \cite{ravi2024sam2segmentimages}). The correspondence errors are a symptom of perhaps a more general limitation: that the policy is upper-bounded by the accuracy of its 3D point inputs. Though policy learning is made simple with 3D points, it does not have information to correct 3D measurement errors.

\textbf{Limitations of triangulation.} We rely on Structure-from-Motion to localize objects over Aria's pre-grasp trajectory. Although this algorithm is less robust when the camera has limited movement, we find that the camera movement from natural task demonstration is usually sufficient. Furthermore, triangulation requires stationary objects, which means that we cannot track objects. In the future, stereo cameras or cheap lidar can remove these constraints and allow closed-loop policy learning in stochastic settings. We hope that depth estimation will become easier with better hardware design.

\textbf{Limitations of hand models.} In this work, we use \cite{pavlakos2023reconstructing} and Aria's hand pose to extract a complete action space, both of which introduce slight inaccuracies. Aria's hand pose does not always predict the same location on the hand and  \cite{pavlakos2023reconstructing}  predicts inconsistently incorrect rotational and translational components on the hand. Even when carefully Equation \ref{eq:action_space} is tuned, the action labels contain 1-2cm error, preventing the policy from solving high-precision tasks. We hope that hand estimation methods will become more reliable with better research and hardware design.

\section{Discussion}
In this work, we presented \textbf{\method{}, a minimal system that trains zero-shot robot policies on in-the-wild egocentric human data without any robot data}. We formalize the morphology-agnostic state-action spaces from prior works and demonstrate how point representations hold the same properties in egocentric in-the-wild settings. Because \method{} optimizes for data collection ergonomics, we also demonstrate how to extract unified state and action representations from human data recorded with the Project Aria smart glasses as the only hardware. As a result, we introduce novel data processing and policy learning design; we demonstrate the importance of each of these components in our baseline and ablation experiments. Although \method{} represents an initial proof-of-concept of how to achieve strong zero-shot transfer from human data, we also acknowledge a handful of limitations, many of which we hope will improve as hardware and robot learning methods improve together.

\textbf{Towards human-centric robotics.} Ultimately, human data carries huge potential in its scalability and morphological completeness. We hope that \method{} will serve as a framework on which future research can extend to fully dexterous and bimanual setups. We hope that our work offers a potentially new theme in robots that is more human-centric, scalable, and abundant.

\clearpage
\acknowledgments{We thank Zhuoran Chen at New York University for their helpful feedback and contribution to the website. We also thank Joyce Kwak for editing the videos. This work was supported by grants from Microsoft, Honda, Hyundai, NSF award 2339096, and ONR award N00014-22-1-2773. LP is supported by the Sloan and Packard Fellowships.}


\bibliography{example}  

\newpage
\appendix
\section{Human Demonstrations}
\label{appendix:tasks}
\begin{figure}[H]
  \centering
  \begin{subfigure}[b]{0.72\textwidth}
    \includegraphics[width=\textwidth]{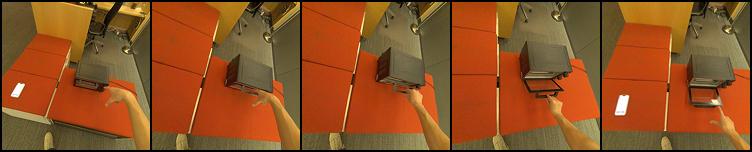}
  \end{subfigure}
  \hfill
  \begin{subfigure}[b]{0.72\textwidth}
    \includegraphics[width=\textwidth]{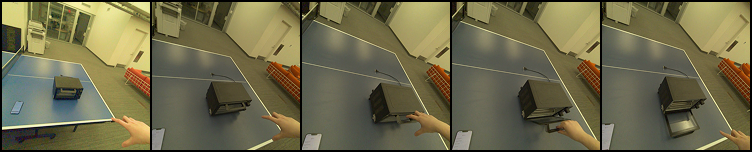}
  \end{subfigure}
  \hfill
  \begin{subfigure}[b]{0.72\textwidth}
    \includegraphics[width=\textwidth]{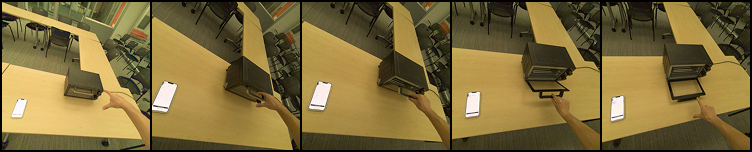}
  \end{subfigure}
  \caption{Open oven door.}
\end{figure}
\begin{figure}[H]
  \centering
  \begin{subfigure}[b]{0.72\textwidth}
    \includegraphics[width=\textwidth]{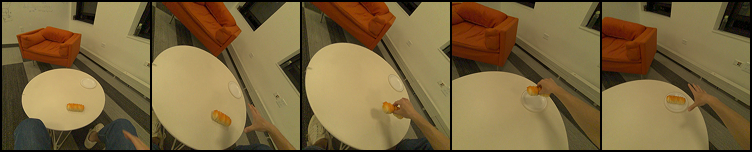}
  \end{subfigure}
  \hfill\begin{subfigure}[b]{0.72\textwidth}
    \includegraphics[width=\textwidth]{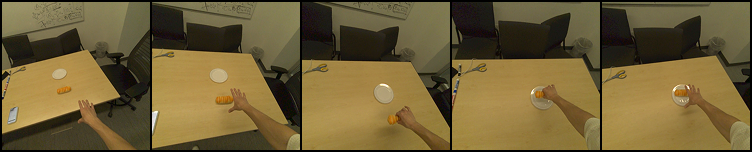}
  \end{subfigure}
  \hfill
  \begin{subfigure}[b]{0.72\textwidth}
    \includegraphics[width=\textwidth]{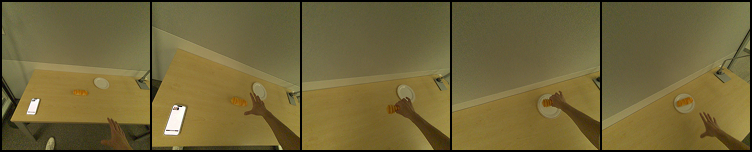}
  \end{subfigure}
  \caption{Put bread on plate.}
\end{figure}
\begin{figure}[H]
  \centering
  \begin{subfigure}[b]{0.72\textwidth}
    \includegraphics[width=\textwidth]{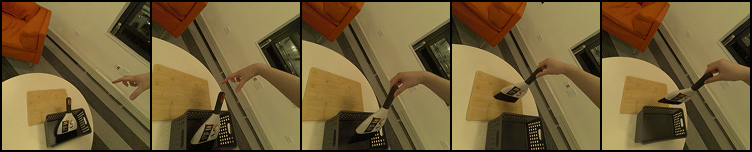}
  \end{subfigure}
  \hfill
  \begin{subfigure}[b]{0.72\textwidth}
    \includegraphics[width=\textwidth]{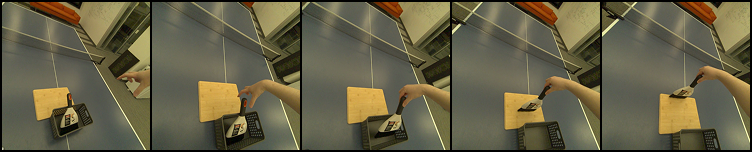}
  \end{subfigure}
  \hfill
  \begin{subfigure}[b]{0.72\textwidth}
    \includegraphics[width=\textwidth]{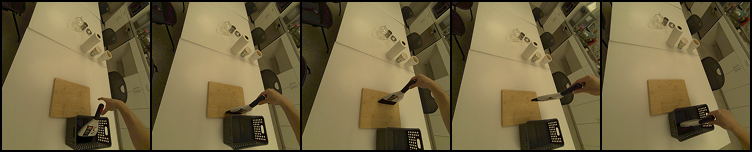}
  \end{subfigure}
  \caption{Sweep board with broom.}
\end{figure}
\begin{figure}[H]
  \centering
  \begin{subfigure}[b]{0.72\textwidth}
    \includegraphics[width=\textwidth]{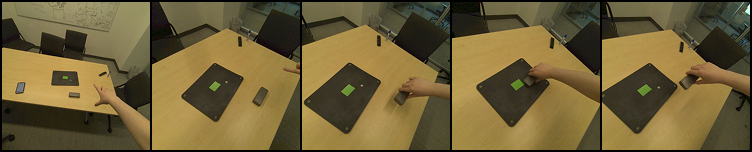}
  \end{subfigure}
  \hfill
  \begin{subfigure}[b]{0.72\textwidth}
    \includegraphics[width=\textwidth]{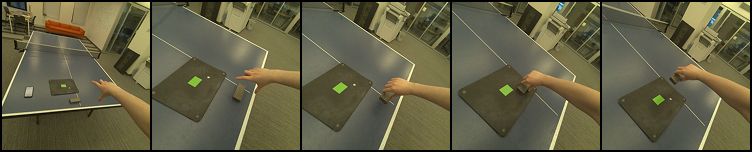}
  \end{subfigure}
  \hfill
  \begin{subfigure}[b]{0.72\textwidth}
    \includegraphics[width=\textwidth]{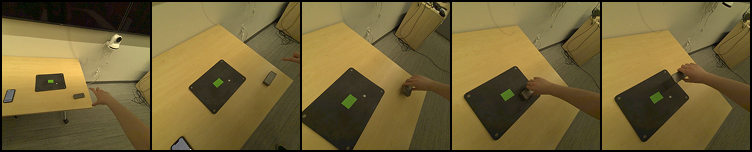}
  \end{subfigure}
  \caption{Erase board.}
\end{figure}
\begin{figure}[H]
  \centering
  \begin{subfigure}[b]{0.72\textwidth}
    \includegraphics[width=\textwidth]{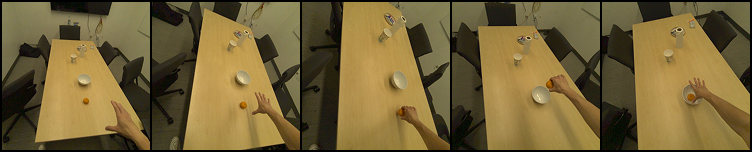}
  \end{subfigure}
  \hfill
  \begin{subfigure}[b]{0.72\textwidth}
    \includegraphics[width=\textwidth]{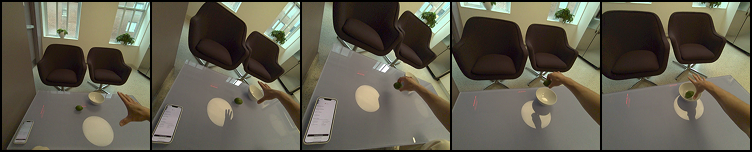}
  \end{subfigure}
  \hfill
  \begin{subfigure}[b]{0.72\textwidth}
    \includegraphics[width=\textwidth]{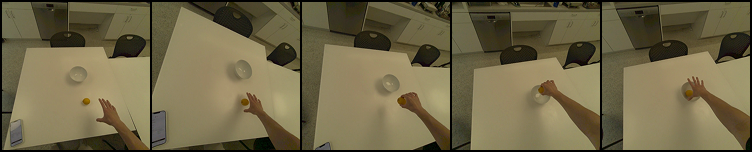}
  \end{subfigure}
  \caption{Sort fruit in bowl.}
\end{figure}
\begin{figure}[H]
  \centering
  \begin{subfigure}[b]{0.72\textwidth}
    \includegraphics[width=\textwidth]{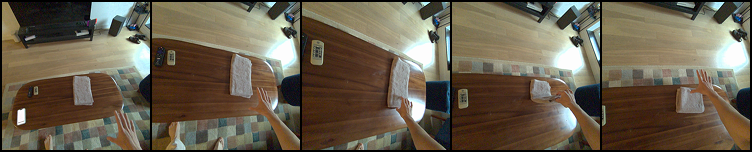}
  \end{subfigure}
  \hfill
  \begin{subfigure}[b]{0.72\textwidth}
    \includegraphics[width=\textwidth]{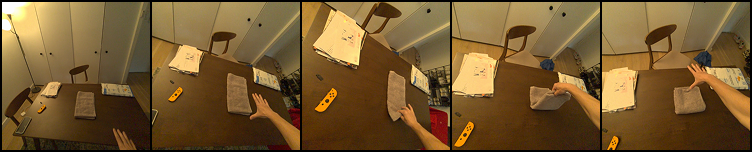}
  \end{subfigure}
  \hfill
  \begin{subfigure}[b]{0.72\textwidth}
    \includegraphics[width=\textwidth]{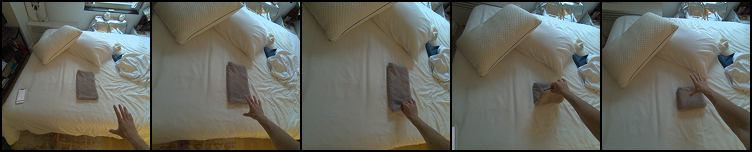}
  \end{subfigure}
  \caption{Fold towel.}
\end{figure}
\begin{figure}[H]
  \centering
  \begin{subfigure}[b]{0.72\textwidth}
    \includegraphics[width=\textwidth]{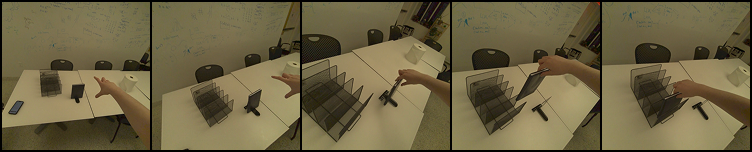}
  \end{subfigure}
  \hfill
  \begin{subfigure}[b]{0.72\textwidth}
    \includegraphics[width=\textwidth]{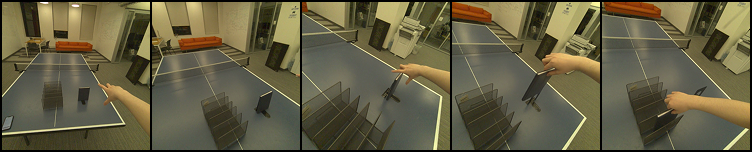}
  \end{subfigure}
  \hfill
  \begin{subfigure}[b]{0.72\textwidth}
    \includegraphics[width=\textwidth]{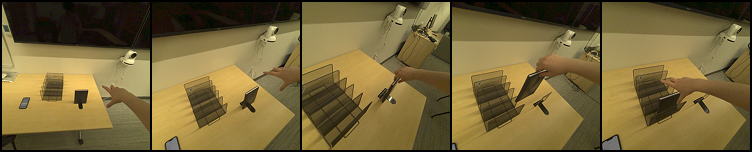}
  \end{subfigure}
  \caption{Insert book in shelf.}
\end{figure}

\section{Triangulating Object Keypoints}\label{appendix:triangulation}

We estimate 3D coordinates $\mathbf{q}^* \in \mathbb{R}^3$ of an object point in the world frame at $t=0$ from 2D observations $\{(T_i, u_i)\}_{i=1}^N$, where $u_i \in \mathbb{R}^2$ is the UV coordinate tracked in frame $i$, and $T_i \in SE(3)$ is the camera-to-world transformation at frame $i$. Let $K$ denote the camera intrinsics and $P_i = K [R_i \mid \mathbf{t}_i] = K T^{-1}_i$ denote the projection matrix from world to image space at frame $i$.

\paragraph{1. Epipolar Filtering.}
To discard geometrically inconsistent views, we apply pairwise epipolar constraints. Given two frames $i$ and $j$, we compute the fundamental matrix:

\begin{equation}
    F_{ij} = K^{-T}[\mathbf{t}_{ij}]_\times R_{ij} K^{-1},
\end{equation}

where $R_{ij} = R_j R_i^\top$, $\mathbf{t}_{ij} = \mathbf{t}_j - R_{ij} \mathbf{t}_i$, and $[\cdot]_\times$ is the skew-symmetric matrix. A frame $i$ is retained if it satisfies the epipolar constraint with at least $m$ other frames:

\begin{equation}
    \left| \mathbf{u}_j^\top F_{ij} \mathbf{u}_i \right| < \epsilon \quad \text{for at least } m \text{ views}.
\end{equation}

\paragraph{2. Robust RANSAC Triangulation.}
Using the filtered inlier views, we perform RANSAC over subsets of size $k$ to find the best triangulated candidate $\mathbf{q}^*$ minimizing reprojection error:

\begin{equation}
    \mathbf{q}_{\text{RANSAC}} = \arg\min_{\mathbf{q}} \sum_{i \in \mathcal{I}} \mathbb{1}\left(\left\| u_i - \mathcal{P}(T_i^{-1} \mathbf{q}) \right\|_2 < \tau\right).
\end{equation}

\paragraph{3. Least Squares with Depth Bias.}
We refine $\mathbf{q}_{\text{RANSAC}}$ via nonlinear least squares with a Huber loss and a soft depth penalty:

\begin{equation}
    \mathbf{q}^* = \arg\min_{\mathbf{q} \in \Omega} \sum_{i \in \mathcal{I}} \left\| u_i - \mathcal{P}(T_i^{-1} \mathbf{q}) \right\|_\rho + \lambda \mathbf{q}_z,
\end{equation}

where $||\cdot||_\rho$ is the Huber loss, $q_z$ is the depth (z-coordinate in world frame), $\lambda$ is the depth bias coefficient, and $\Omega = [\mathbf{l}, \mathbf{u}]$ is a bounding box constraint (i.e. $\mathbf{q}_z > 0$). This formulation encourages geometrically consistent triangulation while avoiding ambiguous far-away solutions in cases of degenerate motion or lag in Cotracker3 predictions.

\paragraph{4. Unified Object Representations.} We repeat Steps 1-3 for each point that we label on the object, and concatenate each triangulated object point to obtain the object representation for the entire trajectory $\tilde{s}$.

\section{Policy Inference}

\begin{algorithm}[H]
\caption{\method{} Policy Inference}
\begin{algorithmic}[1]
\STATE Obtain object keypoints on first frame using DIFT 
\cite{tang2023emergentcorrespondenceimagediffusion} on annotated dataset frame
\STATE Initialize $\tilde{s} = []$
\FOR{$u$ in DIFT labels}
    \STATE Read depth at $u$ from iPhone
    \STATE Unproject $u$ with depth into egocentric frame to obtain $x_u$
    \STATE $\tilde{s} \leftarrow [\tilde{s}, x_u]$
\ENDFOR

\STATE Initialize robot state $\tilde{a}_0$ and history buffer $H = [\tilde{a}_0, ..., \tilde{a}_0]$ of length $h$
\FOR{$t$ in rollout}
    \STATE Compute action chunk $(\tilde{a}_t, ..., \tilde{a}_{t+\ell}) \sim \pi(\tilde{s}_t, H)$ and apply temporal aggregation to get $\tilde{a}_t$
    \STATE Parse gripper action $g \leftarrow \text{bool}(\tilde{a}_t > 0)$
    \STATE Execute $[\tilde{a}_t, g]$ on robot
    \STATE Update buffer $H \leftarrow [H, \tilde{a}_t]_{-h}$
\ENDFOR
\end{algorithmic}
\end{algorithm}

\section{Monocular Depth Estimation}
\label{appendix:monocular}

We record a walkaround of 5 Aruco tags on the table from the Aria glasses and fit an affine scale/shift that minimizes the residual of the depth map at these Aruco tags. Even after calibration, we see that the depth signal deviates with variance from the ground truth Aruco detection, suggesting that monocular depth models are potentially spatio-temporally warped.

\begin{figure}[h]
  \centering
  \begin{subfigure}[b]{1\textwidth}
    \includegraphics[width=\textwidth]{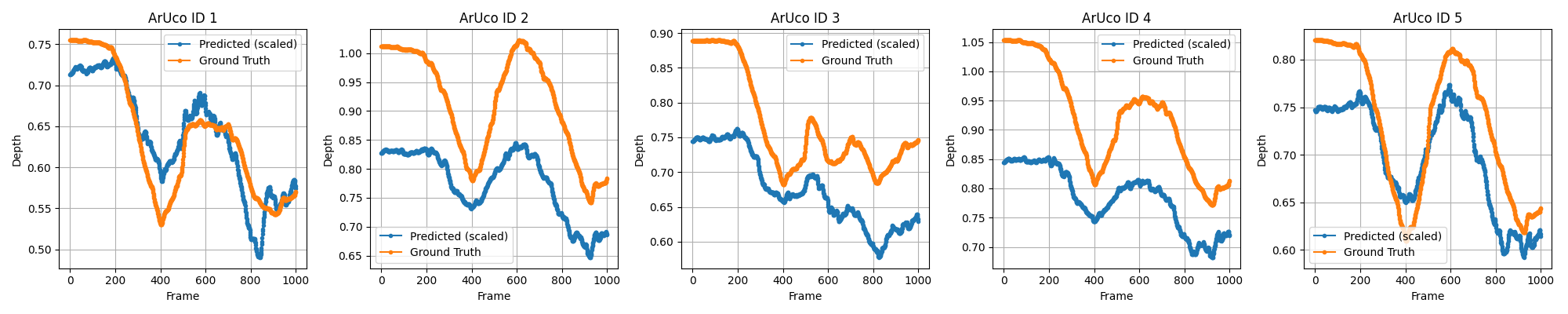}
  \end{subfigure}
  \caption{Monocular depth estimation \cite{Bochkovskii2024:arxiv} calibrated to Aruco tags in the scene.}
\end{figure}

\end{document}